\documentclass{article}
\pdfoutput=1

\PassOptionsToPackage{numbers, compress}{natbib}

\usepackage[preprint]{neurips_2024}

\usepackage{amsmath}
\usepackage[utf8]{inputenc} 
\usepackage[T1]{fontenc}    
\usepackage{hyperref}       
\usepackage{url}            
\usepackage{booktabs}       
\usepackage{amsfonts}       
\usepackage{nicefrac}       
\usepackage{microtype}      
\usepackage{xcolor}         
\usepackage{natbib}
\usepackage{graphicx}
\usepackage[capitalize]{cleveref}
\crefname{section}{Sec.}{Secs.}
\Crefname{section}{Section}{Sections}
\crefname{figure}{Figure.}{Figures.}
\Crefname{figure}{Figure}{Figures}
\crefname{table}{Table.}{Tables.}
\Crefname{table}{Table}{Tables}
\usepackage{adjustbox}
\usepackage{diagbox}
\usepackage{multirow}
\usepackage{makecell}
\usepackage{subfigure}
\usepackage{amssymb}
\usepackage{wrapfig}
\usepackage{colortbl}

\title{FactorLLM: Factorizing Knowledge via Mixture of Experts for Large Language Models}

\makeatletter
\def\thanks#1{\protected@xdef\@thanks{\@thanks
        \protect\footnotetext{#1}}}
\makeatother

\begin{document}

\setlength{\parskip}{0.5em} 

\newcommand*\samethanks[1][\value{footnote}]{\footnotemark[#1]}
\DeclareUrlCommand\url{\color{magenta}}

\author{
Zhongyu Zhao$^{1,}$\footnotemark[1] \quad 
Menghang Dong$^{1,}$\footnotemark[1] \quad 
Rongyu Zhang $^{1,}$\footnotemark[1] \quad
Wenzhao Zheng$^{2,}$\footnotemark[1] $^ ,$\footnotemark[2] \\ \quad 
\textbf{Yunpeng Zhang}$^{3}$\quad
\textbf{Huanrui Yang}$^{4}$\quad 
\textbf{Dalong Du}$^{3}$\quad 
\textbf{Kurt Keutzer}$^{2}$ \quad 
\textbf{Shanghang Zhang}$^{1,}$\footnotemark[3] \\
$^1$Peking University \quad\quad $^2$UC Berkeley \quad\quad $^3$ PhiGent Robotics \quad\quad $^4$ University of Arizona\\
\texttt{zhaozhongyu2000@pku.edu.cn; wenzhao.zheng@outlook.com }
}

\renewcommand{\thefootnote}{\fnsymbol{footnote}}
\footnotetext[1]{Equal contribution. $\dagger$Project leader. $\ddagger$Corresponding author.
}
\renewcommand{\thefootnote}{\arabic{footnote}}

\maketitle

\begin{abstract}

Recent research has demonstrated that Feed-Forward Networks (FFNs) in Large Language Models (LLMs) play a pivotal role in storing diverse linguistic and factual knowledge. Conventional methods frequently face challenges due to knowledge confusion stemming from their monolithic and redundant architectures, which calls for more efficient solutions with minimal computational overhead, particularly for LLMs. In this paper, we explore the FFN computation paradigm in LLMs and introduce \textit{FactorLLM}, a novel approach that decomposes well-trained dense FFNs into sparse sub-networks without requiring any further modifications, while maintaining the same level of performance. Furthermore, we embed a router from the Mixture-of-Experts (MoE), combined with our devised Prior-Approximate (PA) loss term that facilitates 
the dynamic activation of experts and knowledge adaptation, thereby accelerating computational processes and enhancing performance using minimal training data and fine-tuning steps. \textit{FactorLLM} thus enables efficient knowledge factorization and activates select groups of experts specifically tailored to designated tasks, emulating the interactive functional segmentation of the human brain. Extensive experiments across various benchmarks demonstrate the effectiveness of our proposed \textit{FactorLLM} which achieves comparable performance to the source model securing up to 85\% model performance while obtaining over a 30\% increase in inference speed. Code: \url{https://github.com/zhenwuweihe/FactorLLM}.

\end{abstract}

\section{Introduction}
\label{sec:intro}

Large language models\cite{zhao2023llmsurvey,minaee2024llmsurvey} (LLMs) exhibit exceptional capabilities in knowledge recall, attributable to both their extensive training on expansive text corpora\cite{raffel2019c4,gao2020pile,li2023starcoder,cerebras2023slimpajama,xu2024corporasurvey} and their advanced cascade transformer architectures. Central to these architectures are the feed-forward layers within the transformer blocks. These layers constitute a significant fraction of the model's parameters and play a crucial role in storing and processing vast quantities of information\cite{touvron2023llama,touvron2023llama2,Radford2018gpt1,Radford2019gpt2,brown2020gpt3}. However, the substantial size and complexity of transformers primarily stem from their monolithic Feed-Forward Networks\cite{geva2021ffn,dai2022ffn}, which leads to oversized knowledge storage for specific tasks and significant consumption of time and computational resources\cite{meta2024llama3,brown2020gpt3,Bai2023qwen}. These inefficiencies present substantial challenges in efficiently deploying large language models, particularly in computational-constraint task-specific scenarios. Redundant parameters often result in ineffective computations and increase the likelihood of an "illusion" caused by knowledge that is irrelevant to certain tasks. 


Substantial studies \cite{tay2022efficient-transformer,li2022pst,mirzadeh2023relu,liu23dejavu,wan2023efficientllmsurvey} have targeted improvements in the efficiency and adaptability of LLMs. Small Language Models \cite{schick2021slm,black2021gpt-neo,mitra2023orca,microsoft2023phi2,zhang2024tinyllama} (SLMs) aim to reduce the demand for computational resources through compact architectures. However, they are historically and empirically constrained by the Scaling Law \cite{kaplan2020scaling}, which leads to significant model degradation. On the other hand, some approaches \cite{wan2023efficientllmsurvey,zhu2023compressionsurvey} enhance and compress LLMs using techniques such as pruning \cite{frantar2023sparsegpt,sun2023wanda,ashkboos2024slicegpt,zhang2024dynamic-sparse,zhang2024pnp} and quantization \cite{frantar2022gpt-q,shao2023OmniQuant,liu2024qllm,liu2024intactkv}. Moreover, Parameter-Efficient Fine Tuning \cite{han2024peftsurvey,houlsby2019petl-adapter,hu2021lora,dettmers2023qlora} (PEFT) adapts LLMs for specific tasks by integrating few extra trainable parameters. However, these strategies typically require a further phase of training with adequate data, which can obstruct the rapid deployment of LLMs under resource-constrained settings. Thus, striking a balance among efficiency, training costs, and model performance presents a significant challenge for the adaptation of LLMs, necessitating the development of both efficient and effective solutions.

\begin{figure}
    \centering
    \includegraphics[width=\textwidth, keepaspectratio]{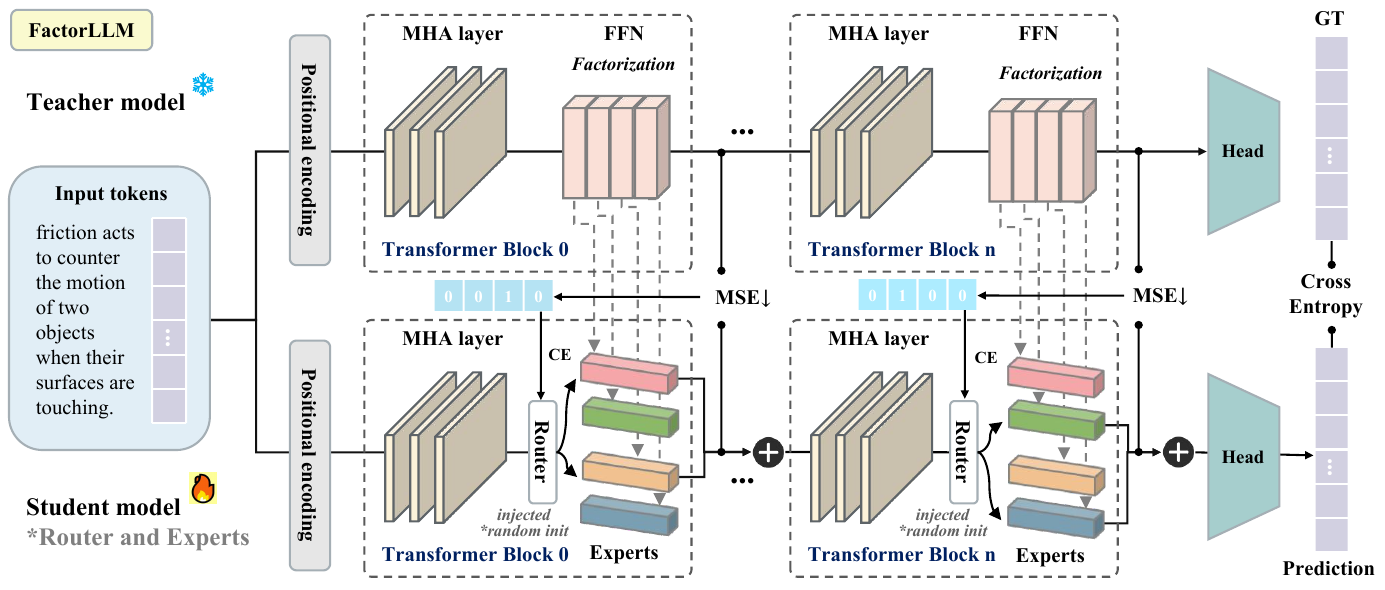}
    \caption{\textbf{Overall Framework of} \textit{FactorLLM}. \textbf{Teacher Model}: Original transformer blocks with multi-head attention (MHA) and feed-forward layers. \textbf{Student Model}: Modified blocks composed of the same MHA layers and factorized FFN, with a linear router deciding which expert(s) tokens will pass through. \textbf{Training Process}: Input tokens branch into normal transformer layers and \textit{FactorLLM} to produce ground-truth (GT) and predictions respectively. Transformers freeze to distill \textit{FactorLLM} based on compositional loss, including mean square error (MSE) between per-layer representations, cross entropy (CE) loss between per-layer optimal and routing masks, and final CE loss between GT and predictions.}
    \label{fig1:framework}
\end{figure}

Drawing on insights from the human brain's capacity to segregate processing for diverse complex tasks \cite{sporns2013network, steele2004segregation, wig2017segregated}, we envisage the densely populated and knowledge-intensive model as a reflection of cerebral mechanisms\cite{zhang2022moefication,zhang2023unimodal}. Therefore, we attempt to segment a \textbf{fully pretrained} and monolithic FFN into multiple subnetworks, each engineered to manage specific types of knowledge, thus enabling knowledge factorization and inference acceleration. In this paper, we introduce \textit{FactorLLM}, an efficient and straightforward framework for decomposing the FFN's weight matrix into modules of identical dimensional shapes that encapsulate task-specific knowledge, as illustrated in \cref{fig1:framework}. This decomposition ensures no loss in performance, as it merely involves reorganizing the matrix elements to their original locations without modifying any values or omitting information.

Given the sparse structure of the decomposed FFN, which is consistent with the Mixture-of-Experts (MoE) architecture \cite{bengio2013deep,shazeer2017outrageously,lepikhin2020gshard, fedus2022switch,du2022glam,zhang2024efficient}, we treat the decomposed subnetworks as experts and leverage the sparse structure to achieve acceleration during inference. Therefore, we integrate a \textbf{randomly initialized} router into each transformer block to enable sparse activation, enhancing the use of specialized knowledge within the experts. However, as the decomposed matrices contain only partial FFN knowledge and the router struggles with reasonable routing, direct fine-tuning of the MoE module can be both inefficient and ineffective. 

Consequently, we utilize a teacher-student framework and implement our devised Prior-Approximate Router (PAR) to expedite knowledge adaptation to different experts for specific tasks. The PAR is introduced with knowledge-similarity conditions by generating pseudo-allocations based on the output features of the student experts and the teacher FFN, thus promoting rapid learning of expert activation strategies rooted in the original model’s prior knowledge. In synergy with PAR, \textit{FactorLLM} boosts computational efficiency during inference by activating a select few experts and facilitates swift adaptation of LLMs to varied knowledge domains using minimal data.
Extensive experiments demonstrate that our proposed \textit{FactorLLM} significantly outperforms baseline decomposition methods by reducing computational overhead by over 30\% while retaining nearly 85\% of the original performance with merely 0.03-0.04\% of training data. The major contribution of our paper can be summarized as follows:
    \begin{itemize}
        \item We propose a simple yet effective approach \textit{FactorLLM} which factorizes the dense FFN in large language models into Mixture of Experts to improve inference efficiency while preserving the original model performance to certain tasks.
        \item We propose Prior-Approximate Router (PAR) by leveraging the existing prior knowledge in LLM that jointly fine-tunes only the injected routers and the factorized experts, facilitating both parameter and data-efficient adaptation of LLMs to specific knowledge domains.
        \item We conducted extensive evaluations to determine the effectiveness and robustness of \textit{FactorLLM} across a variety of model architectures. Our investigations reveal that \textit{FactorLLM} consistently reduces FLOPs by over 30\% while maintaining prediction accuracy above 85\%. 
    \end{itemize}


\section{Related Work}
\label{sec:related}

\subsection{Efficient Large Language Model}
Large language models are frequently criticized for their substantial resource and time demands\cite{samsi2023words-watts,chien2023carbon-impact-gen-ai,DEVRIES2023energy-footprint-ai,akota2024cost-llm} during both training and inference. To address this challenge, various techniques\cite{tay2022efficient-transformer,li2022pst,mirzadeh2023relu,liu23dejavu,wan2023efficientllmsurvey} have been proposed to enhance inference efficiency in large transformer models. Model compression\cite{zhu2023compressionsurvey} is one approach to decrease computational requirements include techniques like pruning\cite{frantar2023sparsegpt,sun2023wanda,ashkboos2024slicegpt,zhang2024dynamic-sparse,zhang2024pnp} and quantization\cite{frantar2022gpt-q,shao2023OmniQuant,liu2024qllm,liu2024intactkv}. Researchers have also developed several resource-efficient and computation-efficient architectures, such as efficient attention mechanisms\cite{zheng2023efficient-attention,zheng2022linear-complexity-attention}, mixture of experts\cite{lepikhin2020gshard,du2022glam,zhang2024efficient}, long-context models\cite{ding2023longnet, ratner2023parallel-context,yen2024longcontext}, and state space models\cite{gu2023mamba}. Additionally, numerous strategies \cite{wan2023efficientllmsurvey} have been identified to improve efficiency throughout the training, fine-tuning, and inference stages. Our objective is to accelerate large language models by modifying their architectures to factorize specific knowledge within the network.


\subsection{Knowledge Decomposition}
Large language models encapsulate extensive knowledge across diverse domains and tasks\cite{hase2021lm-belief}, acquired from vast amounts of training data. To efficiently leverage this knowledge and mitigate architectural redundancy, various methods have been developed to decompose large models and extract intrinsic knowledge. Model editing\cite{zhu2020modifying,sinitsin2020editable,decao2021editing,meng2023locating-editing,meng2023massediting} aims to change the knowledge or brief inside large language models. The Locating-and-Editing\cite{meng2023locating-editing} method views the FFN as a key-value memory\cite{geva2021ffn} and proposes an interpretable approach to trace the effects of weights within the model on the output of input prompts which enables the identification and modification of specific neurons to edit the model's behavior effectively. Alternatively, low-rank matrix decomposition\cite{hsu2022lorf,yuan2023asvd,wang2024svdllm} directly modify model architectures including embedding layer\cite{xu2023tensorgpt} and feed-forward network\cite{li2023losparse} to reallocate knowledge across different modules. Knowledge distillation\cite{hinton2015distilling,hsieh2023distilling,shridhar-etal-2023-distilling,ko2024distillm} is another approach, focusing on transferring knowledge from large models to smaller counterparts. Notably, a novel distillation task termed knowledge factorization\cite{yang2022knowledgefactor} has been proposed to extract both task-agnostic and domain-specific knowledge from neural networks. In our work, we introduce the mixture of experts technique with per-layer distillation training strategy to facilitate effective knowledge factorization.

\subsection{Mixture of Experts}
Mixture of Experts (MoE) \cite{bengio2013deep,shazeer2017outrageously,lepikhin2020gshard, fedus2022switch,du2022glam} is instrumental in integrating diverse domain knowledge across different modules to achieve effective knowledge fusion\cite{wan2024knowledgefusion,zhang2024decomposing}. One method to construct experts involves cloning components of the original transformer block, including attention heads \cite{zhang2022moa}, feed-forward networks \cite{komatsuzaki2023sparseupcycling}, and even bypassing low-rank adapters\cite{luo2024moelora,liu2024intuition}, which has proven to be an effective approach for scaling large transformer-based models and expanding their capacity. Moreover, \cite{zhang2024efficient} situates the traditional MLP in MoE block with the linear-wise feature modulation to further enhance the model efficiency. Alternatively, a different strategy\cite{zhang2022moefication} involves decomposing layers into distinct modules according to K-Means clustering. Our proposed \textit{FactorLLM} extends this concept by adapting the factorized LLM to specific domains of knowledge with a simpler neuron partition, thereby achieving enhanced performance and efficiency.

\section{Proposed Approach}
\label{sec:method}

In this section, we elucidate the rationale behind decomposing the FFN in a fully pretrained LLM into various subnetworks without performance loss and present the comprehensive framework of our proposed \textit{FactorLLM} via MoE. We initially define key concepts and preliminaries concerning LLM and MoE in \cref{sec:pre}. Subsequently, we discuss the factorization of an FFN into multiple subnetworks in \cref{sec:decomp}. Finally, we elaborate on \textit{FactorLLM} with our dynamic routing strategy and the overall training objectives in \cref{sec:factor}. 

\subsection{Preliminary}
\label{sec:pre}
\textbf{Feed-Forward Network (FFN).} For a given input embedding $x \in \mathbb{R}^{d_e}$ and denoting the hidden dimension by $d_h$, the FFN, which are typically implemented as two-layer Multi-Layer Perceptrons (MLP), can be formulated as follows:
\begin{equation}
\label{eq:1}
    \begin{split}
        h&=x\boldsymbol{W}_1+\boldsymbol{b}_1 \\
        F(x)&=\sigma(h)\boldsymbol{W}_2+\boldsymbol{b}_2 
    \end{split}
\end{equation} 
where $F(\cdot)$ stands for the fully connected feed-forward network, $h$ is the hidden representation inside MLP and $\sigma(\cdot)$ is a non-linear activation function (e.g., SiLU\cite{elfwing2017sigmoidweighted}). $\boldsymbol{W}_1\in\mathbb{R}^{d_e\times{d_h}}$ and $\boldsymbol{W}_2\in\mathbb{R}^{d_h\times{d_e}}$ are weight matrices while $\boldsymbol{b}_1\in\mathbb{R}^{d_h}$ and $\boldsymbol{b}_2\in\mathbb{R}^{d_e}$ are bias vectors.



\textbf{Mixture of Experts (MoE).}
The Mixture of Experts model is comprised of a set of $i \in N$ expert functions $E_i(\cdot)$, and a trainable TopK router $R(\cdot)$. The router is designed to distribute input embeddings among the experts by generating a probability vector that dictates the allocation. For a given input embedding $x \in \mathbb{R}^{b\times n\times d_e}$, the output of the MoE model is a composite of contributions from each expert. These contributions are weighted according to the probabilities assigned by the router and can be formally expressed as:
\begin{equation}
\label{equ:2}
    \begin{aligned}
y = \sum_{i=1}^{N}E_{i}(x)R_{i}(x)&, \ \ R(x)=\epsilon(x\boldsymbol{W}_{3}+\boldsymbol{b}_3) \\
s.t. \ \  R(x)\geq0 \ \ &\text{and} \ \ \sum_{i=1}^{N}R_{i}(x)=1
    \end{aligned}
\end{equation}
where $\epsilon(\cdot)$ signifies the softmax function, $\boldsymbol{W}_{3}\in\mathbb{R}^{N\times{d_e}}$ represents a matrix of trainable weights, and $\boldsymbol{b}_{3}\in \mathbb{R}^{N}$ is the bias vector. However, MoE-based architectures often suffer performance degradation when too many inputs are routed to a few experts\cite{fedus2022switch, lepikhin2020gshard}. To mitigate this imbalance, a load balance loss, denoted as $\mathcal{L}_{lb}$, was introduced in \cite{lepikhin2020gshard} to penalize uneven input distribution among experts:
\begin{equation}
\begin{aligned}
\label{equ:Llb}
   \mathcal{L}_{lb} = \frac{K}{N} \sum_{n=1}^{N}\sum_{i=1}^{K}v_{i}(x_{n}) R_{i}(x_n), \quad s.t. \ \ K \leq N
\end{aligned}
\end{equation}
where $x_n$ represents the $n^{th}$ input token. Here, $v_i(x_{n})$ equals 1 if the $i^{th}$ expert is selected for processing $x_{n}$ by the TopK selection function, and 0 otherwise. 

\subsection{Model Decomposition}
\label{sec:decomp}
The fundamental concept of model decomposition involves partitioning neurons in a fully pretrained model that frequently activates concurrently into distinct subnetworks for acceleration. These subnetworks are sparsely activated during the feedforward phase, thereby accelerating the model inference without performance loss. To maintain consistent forward processing speeds and reduce the "bucket effects" associated with differing expert sizes, we decompose the weight matrix into $N$ subnetworks of uniform dimensions. Such partition eliminates delays caused by the slowest component in parallel operations, thereby improving the efficiency of parallel computations.

For the designated fully pretrained FFN $F(\cdot)$, the factorized subnetworks are characterized by weight matrices $\boldsymbol{W}^{(i)}_1 \in \mathbb{R}^{d_e \times d_n} \ \text{and} \ \boldsymbol{W}^{(i)}_2 \in \mathbb{R}^{d_n \times d_e}$, alongside biases $\boldsymbol{b}^{(i)}_1 \in \mathbb{R}^{d_n} \ \text{and} \  \boldsymbol{b}^{(i)}_2 \in \mathbb{R}^{d_e}$. The index $i$ identifies the $i^{th}$ subnetwork in the decomposed FFN, while $d_n = \frac{d_h}{N}$ specifies the hidden layer dimension of each subnetwork. The objective of FFN decomposition is to establish a mapping function $\delta$ that reassigns the original neuron index $p\in\mathbb{R}^{d_h}$ to a new index $q\in\mathbb{R}^{d_h}$. This mapping dictates the configuration of a permutation matrix that reorders the weight matrices into their corresponding subnetworks. Consequently, the permutation matrix $P_\delta\in\mathbb{R}^{d_h\times d_h}$ is defined with the input token $x$ as:
\begin{equation}
\label{eq:3}
(P_\delta)_{pq} =
\begin{cases} 
1, & \text{if} \ \ \delta_q = p  \\
0, & \text{else}
\end{cases}    
\end{equation}

\begin{figure}
    \centering
    \includegraphics[width=\textwidth, keepaspectratio]{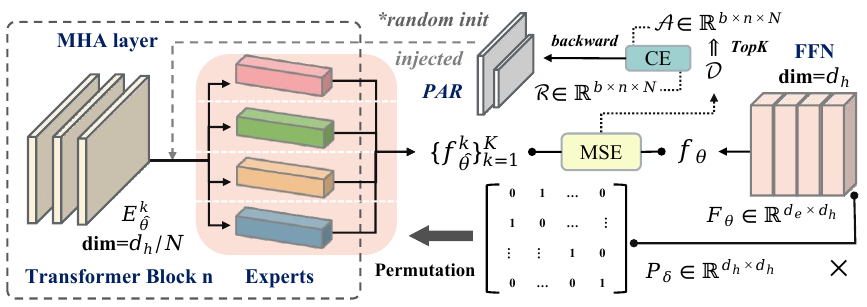}
    \caption{We construct \( N \) experts, \( E^k_{\hat{\theta}} \) (with \(\text{dim} = d_h/N\)), by applying a permutation \( P_\delta \) to the pretrained FFN \( F_\theta \) (with \(\text{dim} = d_h\)) and then dividing it equally. Prior approximate routers (PAR), initially randomly initialized, are placed between the MHA layer and the experts. An MSE loss is computed between the outputs from the FFN block \( f_\theta \) and the best \( K \) experts \( f^k_{\hat{\theta}} \) over a dataset \(\mathcal{D}\). Then, the top \( K \) selections \(\mathcal{A}\) are determined using the TopK algorithm. \(\mathcal{A}\) and the output of the router \(\mathcal{R}\) are combined to compute the cross-entropy (CE) loss.}
    \label{fig:method}
\end{figure}

The subsequent factorization is represented as:
\begin{equation}
\label{eq:4}
\begin{aligned}
\begin{bmatrix}
\boldsymbol{W}^1_1 & \boldsymbol{W}^2_1 & \cdots & \boldsymbol{W}^n_1
\end{bmatrix} &= \boldsymbol{W}_1 P_\delta, \\
\begin{bmatrix}
(\boldsymbol{b}^1_1)^T & (\boldsymbol{b}^2_1)^T & \cdots & (\boldsymbol{b}^n_1)^T
\end{bmatrix} &= \boldsymbol{b}_1 P_\delta, \\
\begin{bmatrix}
(\boldsymbol{W}^1_2)^T & (\boldsymbol{W}^2_2)^T & \cdots & (\boldsymbol{W}^n_2)^T
\end{bmatrix} &= (P_\delta^T \boldsymbol{W}_2)^T.
\end{aligned}
\end{equation}
Note that $\boldsymbol{b}_2$ remains unchanged. Therefore, we can utilize the derivation in \cref{eq:4} to obtain the same expression in \cref{eq:1}:
\begin{equation}
    \begin{split}
        h^\prime=x\boldsymbol{W}_1 P_\delta&+\boldsymbol{b}_1 P_\delta=hP_\delta \\
        S(x)=\sigma(h^\prime)P_\delta^T \boldsymbol{W}_2+\boldsymbol{b}_2&=\sigma(h)P_\delta P_\delta^T \boldsymbol{W}_2+\boldsymbol{b}_2=F(x)    
    \end{split}
\end{equation}
where $S(\cdot)$ represents the groups of factorized subnetworks. Consequently, the monolithic FFN can be decomposed into $N$ subnetworks while preserving the integrity of the output representation processed by the FFN layer.

\subsection{FactorLLM} 
\label{sec:factor}
\textbf{Transforming into Mixtur-of-Experts.} In \cref{sec:decomp}, it is established that partitioning the FFN into $N$ distinct subnetworks does not compromise the overall model efficacy. This finding allows us to exploit the resulting sparse architecture to expedite model computations by selectively activating a limited subset of subnetworks, denoted as $S=\{s_{k}\}_{k=1}^{N}$ . Drawing parallels to the architecture of Mixture of Experts (MoE), we treat these subnetworks $S$ as individual experts $E$:
\begin{equation}
    E(x)=\sum_{i\in{\mathbf{S}}}\sigma(x\boldsymbol{W}^i_1+\boldsymbol{b}^i_1)\boldsymbol{W}^i_2+\boldsymbol{b}_2
\end{equation}
It should be noted that when all experts are activated, i.e., $K=N$, the output of the expert ensemble $E$ is equivalent to the monolithic FFN function $F(x)$.
Moreover, to enhance computational efficiency during the inference phase, a randomly initialized trainable router $R$ is introduced for dynamically activating only $K$ experts, where $K \leq N$. Thus, we factorize the fully pretrained dense LLM into the sparse MoE-LLM together with a randomly initialized injected router, which is designed to facilitate adaptive and efficient computation.

\textbf{Prior Approximate Router (PAR).} 
Given the random initialization of the newly injected router $R$, its initial efficacy in selecting suitable experts for different input tokens $x$ is constrained. To address this, we harness the well-pretraining the original model $\theta$ and design a Prior Approximate Router (PAR) within the factorized model $\hat{\theta}$. PAR, embedded in a teacher-student framework, aims to minimize discrepancies in expert selection with a tailored PA loss term, steering $R$ towards experts $E_{\hat{\theta}}$ whose knowledge aligns closely with the teacher $F_\theta$.

We define $f_{\theta}$ as the output of the teacher's feed-forward network $F_{\theta}(x)$, and $\{f^{k}_{\hat{\theta}}\}_{k=1}^{K}$ as the outputs from the student experts $E_{\hat{\theta}}(x)$. We first compute the Mean Squared Error (MSE) across these features, yielding a set of distances $\mathcal{D}=\{d^{mse}_{k}\}_{k=1}^{K}$. Subsequently, we apply the TopK algorithm to extract expert indices $\mathcal{I}$ for the smallest $d^{mse}$, leading to a pseudo router allocation $\mathcal{A} \in \mathbb{R}^{b \times n \times N}$, where elements corresponding to indices in $\mathcal{I}$ are set to 1 and all others to 0, defined as $\mathcal{A} = index(TopK(\mathcal{D}))$. Therefore, leveraging the pre-established pseudo label $\mathcal{A}$, we expedite the router's update using the cross-entropy function:
\begin{equation}
\mathcal{L}_{PA} = -\frac{1}{N} \sum_{l=1}^{L} \sum_{n=1}^{N} \mathcal{A}_{n}^{l} \log \mathcal{R}_{n}^{l}
\end{equation}
Here, $L$ denotes the number of layers in $\hat{\theta}$, and $\mathcal{R} \in \mathbb{R}^{b \times n \times N}$ represents the router's output within $\hat{\theta}$.
\textbf{Optimization.}
Our proposed \textit{FactorLLM} employs a teacher-student framework, transferring knowledge from the teacher FFN to a newly initialized router and utilizing ground truth to update both the router and the experts concurrently. Given data samples $\mathcal{X}$ and corresponding labels $\mathcal{Y}$, the model's predictions are represented by $\mathcal{F}=\hat{\theta}(\mathcal{X})$, and the finetuning loss is expressed as:
\begin{equation}
\mathcal{L}_{FT} = -\frac{1}{M} \sum_{j=1}^{M} \sum_{c=1}^{C} \mathcal{Y}_{j}^{c} \log \mathcal{F}_{j}^{c}
\end{equation}
where $M$ represents the total number of samples, $C$ represents the size of the vocabulary. To promote task-specific knowledge adaptation, we omit the balance loss during finetuning and integrate our custom PA loss, leading to the comprehensive optimization objective:
\begin{equation}
\mathcal{L}_{overall} = \mathcal{L}_{FT} + \alpha \times \mathcal{L}_{PA}
\end{equation}
where $\alpha$ is a hyperparameter that balances model generalization and expert specialization.

\section{Experiments}
\label{sec:experiment}
In this section, we will first describe the experimental setup, methodologies, and evaluation metrics used to assess the performance of our proposed language model in \cref{sec:experiment}. Subsequently, we present quantitative results of \textit{FactorLLM} in \cref{sec:performance} and analyze efficiency of our method in \cref{sec:efficiency}. Finally, \cref{sec:ablation} shows ablation studies conducted to demonstrate the effectiveness of method designs.
\subsection{Experiment Setup}
\label{sec:experiment}
\textbf{Implementation Details.}
We utilize the TinyLlama\cite{zhang2024tinyllama} with 1.1 billion parameters and MobileLlama\cite{chu2024mobilevlm} with 1.4 billion parameters as the backbones for our Large Language Models (LLMs). Optimization is carried out using the Adam algorithm\cite{kingma2017adam}, configured with a learning rate of 4e-5 and $(\beta_1, \beta_2) = (0.9, 0.95)$. The settings include a weight decay of 1e-5 and a gradient clipping threshold of 1. We set the batch size to 64 and the sequence length to 1024. 

The warmup stage is conducted before training FactorLLM to guide our PAR to master ability to choose experts and then routers will be frozen to train experts in every layer. In our experiments, training was limited to 100,000 steps, corresponding to approximately 0.03\% of the data used to train the original model. These experiments were conducted on a single NVIDIA GeForce RTX 4090 equipped with 24GB GDDR6X VRAM.

\textbf{Datasets and Baselines.} We train our model using the Pajama dataset. The dataset we used was approximately 0.05\% of the total Pajama dataset and we split them into the training and development sets with the ratio of 99:1. To evaluate the performance of our model, we used several natural language understanding datasets, including HellaSwag\cite{Zellers2019HellaSwagCA}, OpenBookQA\cite{Mihaylov2018openbookqa}, Winogrande\cite{sakaguchi2019winogrande}, ARC-Easy\cite{Clark2018ARC}, ARC-Challenge\cite{Clark2018ARC}, BoolQ\cite{clark2019boolq}, PIQA\cite{bisk2019piqa} and MMLU\cite{hendryckstest2021mmlu}.

We benchmark our proposed FactorLLM against two state-of-the-art baselines: MoEfication\cite{zhang2022moefication}, which decomposes the FFN into MoE using the K-Means algorithm to initialize and construct experts, and KnowledgeFactor\cite{yang2022knowledgefactor} that decomposes model knowledge by separating into common knowledge and task-specific subnets. Evaluations are based on two widely-used LLM backbones, TinyLlama\cite{zhang2024tinyllama} and MobileLlama\cite{chu2024mobilevlm}. MoEfication, KnowledgeFactor and FactorLLM restrict fine-tuning to the factorized MoE blocks. Our model's variants are distinguished by two parameters: the number of routers ($R$), the total number of experts ($E$) and the number of experts selected by the router ($K$). For instance, the base variant of FactorLLM, denoted as $1R4E2K$, incorporates 1 router, total 4 experts and 2 experts are selected to process each token.

\subsection{Quantitative Performance}
\label{sec:performance}
In this section, we present a comprehensive analysis of the quantitative performance of \textit{FactorLLM} configured with $1R$ and $4E$ across diverse benchmarks on two distinct LLM platforms, TinyLlama\cite{zhang2024tinyllama} and MobileLlama\cite{chu2024mobilevlm}. We further explore configurations involving varying numbers of experts, ranging from one to three.

As indicated in \cref{tab:evaluation_scores}, the evaluation results illustrate that \textit{FactorLLM} consistently delivers considerable enhancements in inference efficiency while sustaining robust accuracy levels. Our findings reveal that \textit{FactorLLM} surpasses both MoEfication\cite{zhang2022moefication} and KnowledgeFactor\cite{yang2022knowledgefactor} across various expert activations facilitated by our novel Prior-Approximate Router. Notably, for the boolq dataset, \textit{FactorLLM-$3K$} surpasses the established upper bounds by directly fine-tuning on TinyLlama and MobileLlama by margins of 3.9\% and 1.7\%, respectively. Remarkably, even in its most efficient configuration—activating a singular expert—\textit{FactorLLM} still exceeds MoEfication's performance on datasets such as openbookqa, hellaswag, and arc$\_$e by a significant margin of over 0.03. These results from our detailed performance evaluation affirm that \textit{FactorLLM} not only conserves but often amplifies the accuracy of large language models while markedly diminishing computational demands.

\begin{table*}
    \caption{\textbf{Performance evaluation.} We assess prediction accuracy (\%) using TinyLlama\cite{zhang2024tinyllama} and MobileLlama\cite{chu2024mobilevlm}, comparing FactorLLM of different settings against other baselines.}
    \label{tab:evaluation_scores}  
    \centering
    \setlength\tabcolsep{5pt} 
    \begin{adjustbox}{width=\linewidth}  
        \begin{tabular}{c|c|cccccccc}  
        \toprule
        Backbone &Method & winogrande & piqa & openbookqa & mmlu & hellaswag & boolq & arc\_e & arc\_c \\
        \midrule
        \multirow{6}{*}{TinyLlama\cite{zhang2024tinyllama}} & Upper bound & 59.1 & 73.2 & 36.0 & 24.5 & 59.2 & 57.8 & 55.3 & 30.1 \\
        \cmidrule{2-10}
        & MoEfication\cite{zhang2022moefication} & 53.7 & 55.1 & 23.6 & 23.0 & 27.9 & 56.1 & 29.9 & 22.6 \\
        & KnowledgeFactor\cite{zhang2022moefication} & 51.8 & 62.4 & 27.6 & 24.0 & 39.0 & 58.1 & 41.3 & 24.2 \\
        & \cellcolor{green!10}FactorLLM-$1K$  & \cellcolor{green!10}51.1 & \cellcolor{green!10}57.2 & \cellcolor{green!10}25.6 & \cellcolor{green!10}22.9 & \cellcolor{green!10}30.8 & \cellcolor{green!10}54.9 & \cellcolor{green!10}33.3 & \cellcolor{green!10}23.5 \\
        & \cellcolor{green!10}FactorLLM-$2K$  & \cellcolor{green!10}53.1 & \cellcolor{green!10}63.3 & \cellcolor{green!10}30.4 &\cellcolor{green!10} 24.1 & \cellcolor{green!10}39.6 & \cellcolor{green!10}57.2 & \cellcolor{green!10}41.7 & \cellcolor{green!10}24.1 \\
        & \cellcolor{green!10}FactorLLM-$3K$  & \cellcolor{green!10}55.9 & \cellcolor{green!10}69.2 & \cellcolor{green!10}31.8 & \cellcolor{green!10}24.2 & \cellcolor{green!10}49.5 & \cellcolor{green!10}61.7 & \cellcolor{green!10}50.6 & \cellcolor{green!10}26.5 \\
        \midrule
        \multirow{6}{*}{MobileLlama\cite{chu2024mobilevlm}}& Upper bound & 58.0 & 72.4 & 34.6 & 24.9 & 55.9 & 57.8 & 61.3 & 28.7 \\
        \cmidrule{2-10}
        &MoEfication\cite{zhang2022moefication} & 52.3 & 58.4 & 24.7 & 23.3 & 29.9 & 57.2 & 38.4 & 23.2 \\
        &KnowledgeFactor\cite{yang2022knowledgefactor} & 52.8 & 61.6 & 27.6 & 23.0 & 35.9 & 54.6 & 39.4 & 23.3 \\
        &\cellcolor{green!10}FactorLLM-$1K$& \cellcolor{green!10}50.0 & \cellcolor{green!10}56.1 & \cellcolor{green!10}26.2 & \cellcolor{green!10}23.0 & \cellcolor{green!10}30.2 & \cellcolor{green!10}62.2 & \cellcolor{green!10}32.7 & \cellcolor{green!10}22.3 \\
        &\cellcolor{green!10}FactorLLM-$2K$& \cellcolor{green!10}52.0 & \cellcolor{green!10}62.1 & \cellcolor{green!10}29.4 & \cellcolor{green!10}24.4 & \cellcolor{green!10}39.2 & \cellcolor{green!10}59.5 & \cellcolor{green!10}41.1 & \cellcolor{green!10}25.9 \\
        &\cellcolor{green!10}FactorLLM-$3K$& \cellcolor{green!10}52.8 & \cellcolor{green!10}63.6 & \cellcolor{green!10}28.4 & \cellcolor{green!10}24.0 & \cellcolor{green!10}43.8 & \cellcolor{green!10}62.5 & \cellcolor{green!10}46.3 & \cellcolor{green!10}26.7 \\

        \bottomrule
        \end{tabular}
    \end{adjustbox}
\end{table*}

\begin{table*}[t]
\centering
    \caption{\textbf{Results of different MoE settings and router designs.} We here examine FactorLLM with different number of experts, activated modules and routers based on TinyLlama.}
\setlength\tabcolsep{7pt} 
\begin{adjustbox}{width=\linewidth}  
    \begin{tabular}{ccc|ccccccccc}
    \toprule
    \#$R$ & \#$E$ & \#$K$ & winogrande & piqa & openbookqa & mmlu & hellaswag & boolq & arc\_e & arc\_c & Maintenance\\
        \midrule
    \multicolumn{3}{c|}{TinyLlama} & 59.1 & 73.2 & 36.0 & 24.5 & 59.2 & 57.8 & 55.3 & 30.1 & 100\% \\
    \midrule
    1   & 4   & 1   &    51.1   &57.2 &   25.6    & 22.9 &    30.8  & 54.9 & 33.3  & 23.5 & 76.7\% \\
    \rowcolor{green!10}1   & 4   & 2   &   53.1       & 63.3   & 30.4     & 24.1  & 39.6    & 57.2    & 41.7    &   24.1 & 85.0\% \\ 
    1   & 4   & 3   &    55.9   &69.2 &   31.8    & 24.2 &    49.5  & 61.7 & 50.6  &  26.5 & 93.2\% \\
    \midrule
    1   & 8   & 4   &    53.1   &64.1 &   31.2    & 23.2&    39.7  & 57.8 & 43.9  &  25.2 & 86.0\% \\
    1   & 16  & 8   &    52.9   &63.4 &   28.8    & 24.2&    39.6  & 59.2 & 43.7  &  24.7  & 85.6\% \\
    \midrule
    3   & 4   & 2   &    48.7   &57.8 &   26.4    & 23.2 &    29.1  & 59.3 & 31.0  &  22.8 & 76.5\% \\
    \bottomrule
    \end{tabular}
        \end{adjustbox}
    \label{tab:moe-setting}
\end{table*}

\subsection{Efficiency Analysis}
\label{sec:efficiency}

\begin{figure}
    \centering
    \begin{minipage}[b]{0.5\textwidth}
        \centering
        \includegraphics[width=\textwidth]{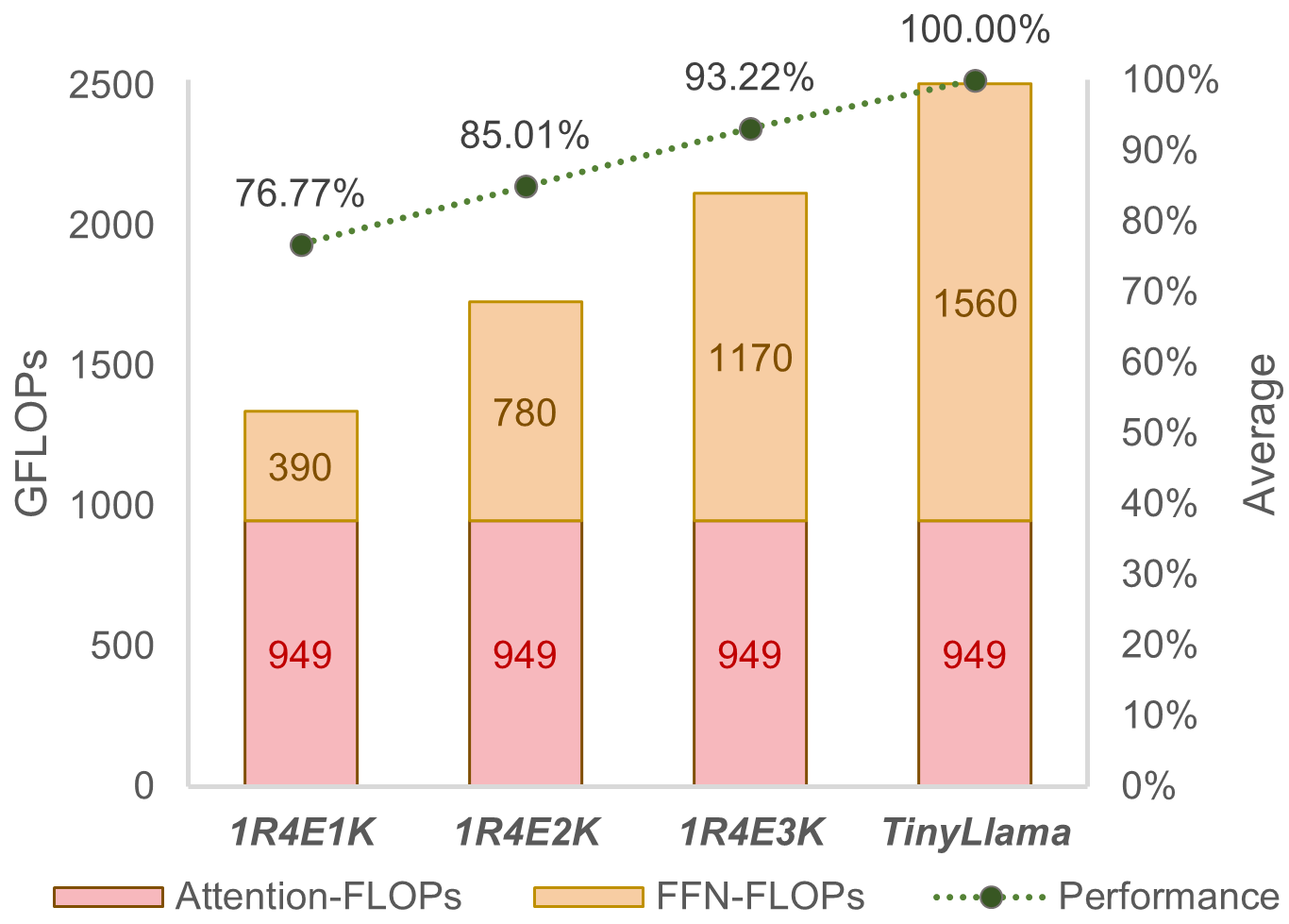}
        \caption{\textbf{Comparison of FLOPs and performance across different model configurations.} The left y-axis represents GFLOPs for both attention and FFN layers, while the right y-axis shows the relative performance percentage.}
        \label{fig:flops}
    \end{minipage}
    \hfill
    \begin{minipage}[b]{0.45\textwidth}
        \centering
        \includegraphics[width=\textwidth]{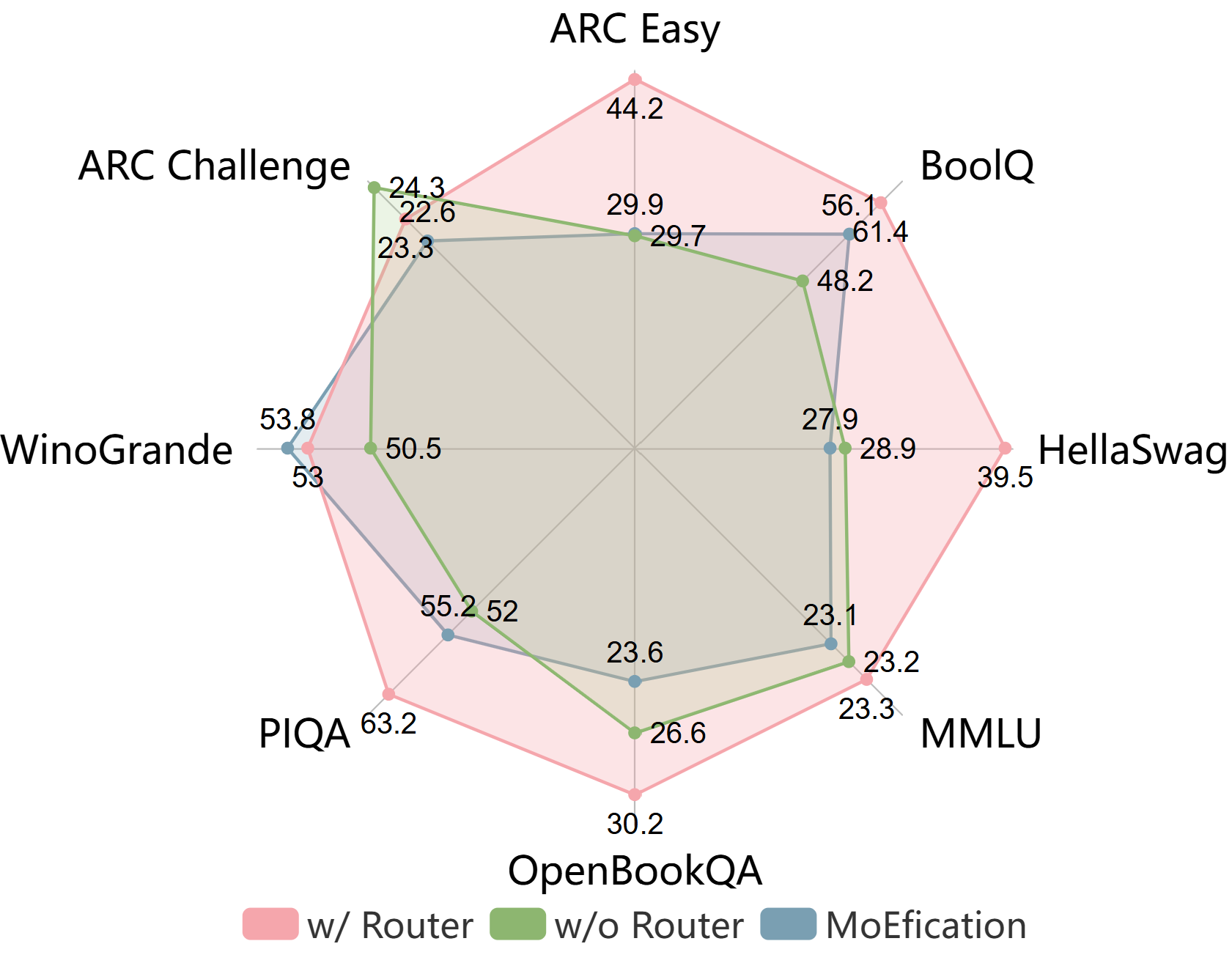}
        \caption{\textbf{Performance comparison between models with and without the router mechanism.} The radar chart highlights differences in multiple performance, demonstrating the impact of router integration.}
        \label{fig:routing}
    \end{minipage}
\end{figure}


\textbf{FLOPs Reduction.} \textit{FactorLLM} markedly reduces inference FFN GFLOPs, as illustrated in \cref{fig:flops}. The extent of FLOPs reduction correlates with the number of activated experts; notably, the $1R4E1K$ configuration achieves the most substantial reduction, approximately 75\%. This efficiency gain stems from the factorization of FFNs into sparser architectures, effectively minimizing computational demands and reducing the overall FLOP count. Additionally, \textit{FactorLLM} can reduce total computational overhead by nearly 50\% under the $1R4E2K$ setup while retaining over 85\% accuracy. However, minimizing parameters in modified feed-forward layers shifts the computational bottleneck to the attention layers, resulting in a drop in accuracy to 76.8\%. The near-linear relationship between GFLOPs and model accuracy underscores the need for future enhancements, particularly in optimizing attention layers.


\textbf{Minimal Data Amount.} \textit{FactorLLM} is designed to adapt quickly to new tasks with minimal data. Our experiments indicate that FactorLLM can maintain over 85\% of the original model performance using merely 0.03-0.04\% of the training data. Specifically, TinyLlama requires 3 trillion tokens to achieve optimal performance while \textit{FactorLLM} can reach our convergence performance levels with just 30M to 50M tokens. This data efficiency is particularly beneficial in scenarios where obtaining large amounts of labeled data is challenging.

\subsection{Routing Analysis}

As shown in \cref{fig:routing}, the model's performance is significantly better when using a router compared to when experts for each layer are randomly selected to process tokens. Specifically, the performance with the router is 85.6\%, while without the router, it is 73.9\%, indicating a difference of approximately 11.7\%. Additionally, \cref{tbl:ablation} shows that if the model is trained without a router from the beginning, the final performance reaches 79.5\%. These data indicate that without a router, the four experts tend to evolve into similar modules during training. In contrast, using a router during training encourages each expert to become more "specific", resulting in experts that are complementary rather than similar, thereby enhancing the model's overall performance.

\begin{wrapfigure}{r}{0.45\textwidth}
    \centering
    \vspace{-1.5em}
    \includegraphics[width=\linewidth]{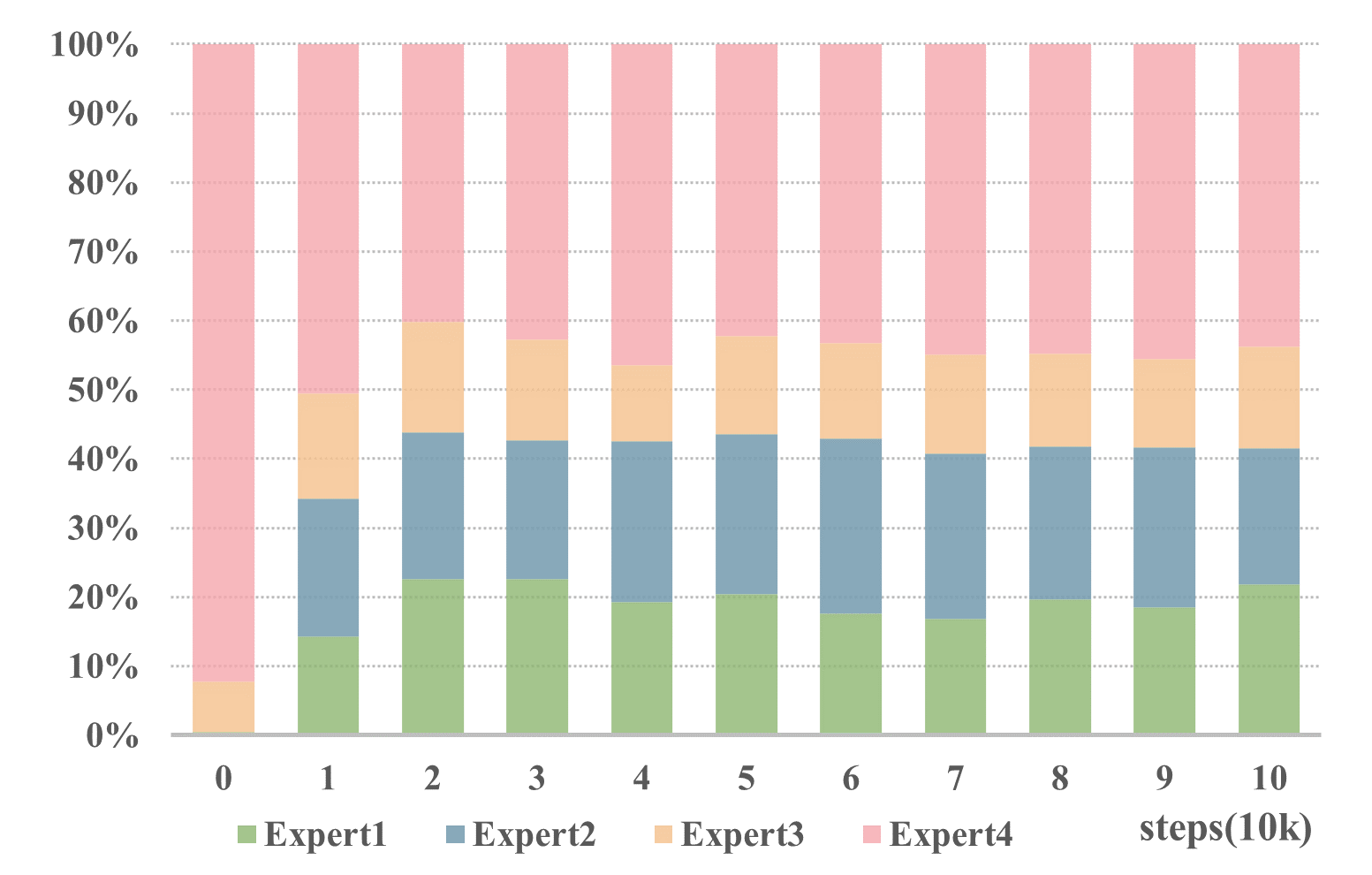}
    \vspace{-2em}
        \caption{Routing patterns $v.s.$ training steps.}
    \vspace{-1.0em}
    \label{fig:tokens}
\end{wrapfigure}

Furthermore, we present the router allocation results on different training steps. As illustrated in \cref{fig:tokens}, the effectiveness of our proposed Prior-Approximate Router (PAR) becomes more apparent as training iterations increase. It is evident that the allocation of experts transitions from chaos to stability, indicating that PAR effectively leverages prior knowledge within the Large Language Model (LLM) to accurately direct task-specific input tokens to the appropriate experts. These observations further validate the efficacy of our proposed PAR.
\par 

\subsection{Ablation Study}
\label{sec:ablation}

We initially assessed the effectiveness of varying configurations of routers ($R$), experts ($E$), and knowledge units ($K$). As depicted in \cref{tab:moe-setting}, maintaining a constant $K/N$ ratio while increasing the number of routers does not improve performance; instead, it may lead to performance deterioration. This suggests that a higher number of experts does not necessarily enhance overall outcomes and might even trigger a decline in performance. This is because a single router more efficiently integrates three experts into a unified expert group. Furthermore, introducing additional routers to better align with the TinyLLama architecture results in performance degradation, achieving only 76.5\% accuracy. We hypothesize that this is due to routing conflicts among the routers during the process of knowledge adaptation from the original LLM. These findings reinforce the efficacy and universality of our \textit{FactorLLM}.

\begin{table*}[t]
\centering
    \caption{\textbf{Ablation study.} This section explores the effectiveness of each component in FactorLLM. Factor refers to the factorization and direct fine-tuning of the Feed-Forward Network (FFN), while PAR denotes the integration of a Prior-Approximate Router.}
\setlength\tabcolsep{6pt} 
\begin{adjustbox}{width=\linewidth}  
    \begin{tabular}{ccc|ccccccccc}
    \toprule
      & Factor & PAR & winogrande & piqa & openbookqa & mmlu & hellaswag & boolq & arc\_e & arc\_c & Maintenance \\ 
              \midrule
    \multicolumn{3}{c|}{TinyLlama} & 59.1 & 73.2 & 36.0 & 24.5 & 59.2 & 57.8 & 55.3 & 30.1 & 100\% \\
    \midrule
    $Ex_0$ & - & - &    49.9    &55.4&   28.6    &23.4&   35.6  &56.9 & 35.2 & 23.9 & 79.5\%  \\
    $Ex_1$ & \checkmark & -    &  50.2      &  58.3   &   26.4     &    23.3    &   34.8        &   57.3    &   37.2    &   23.6 & 79.5\%    \\
    $Ex_2$ & - & \checkmark      &   \textbf{53.3}         &  62.1    &    27.6        &  23.0    &     38.9  &  \textbf{59.5}  &   41.7     &  23.7  & 83.5\%    \\
    \midrule
    \rowcolor{green!10}$Ex_3$ & \checkmark & \checkmark  &   53.1 & \textbf{63.3} & \textbf{30.4} & \textbf{24.1}  & \textbf{39..6} &59.2 & \textbf{41.7}  & \textbf{24.1} & \textbf{85.4\%}\\
    \bottomrule
    \end{tabular}
    \end{adjustbox}
    \label{tbl:ablation}
\end{table*}

As depicted in \cref{tbl:ablation}, a comparison between the first two rows and the last two rows reveals minimal performance variation between these experimental sets. Hence, randomly assigning weights to experts during initialization ($Ex_0$) does not significantly alter the overall performance of 79.5\% accuracy maintenance compared to direct expert splitting ($Ex_1$). This indicates that the initialization strategy and whether to use pretrained models do not substantially impact the ultimate performance. Nonetheless, a discernible contrast exists between random expert selection and Prior-Approximate Router (PAR) to select experts ($Ex_2$), particularly evidenced by the outcomes on the PIQA and ARC-Challenge datasets. This underscores the superiority of our proposed PAR. Additionally, when integrating these approaches ($Ex_3$), \textit{FactorLLM} achieves an optimal performance level of 85.4\%.

\section{Conclusion and Limitations}
\label{sec:conclusion}
In this paper, we introduce FactorLLM, an efficient and streamlined framework designed to enhance inference speed and facilitate rapid adaptation of LLMs to task-specific knowledge using the proposed Prior-Approximate Router (PAR). FactorLLM factorizes the FFN weight matrix into modules of uniform dimensional shapes that store task-specific knowledge and preserves original performance integrity by avoiding any modification or omission of data. Furthermore, it enables fine-tuning to specific tasks with minimal data requirements. Although FactorLLM has demonstrated promising outcomes, there is considerable potential for enhancing its performance. In future work, we aim to develop advanced strategies for parameter partitioning, potentially segmenting parameters into experts of varying shapes to better address diverse tasks. This approach could further optimize the architecture and improve the model's adaptability to specific requirements.



{
\small

\bibliography{references}
}

\end{document}